\pdfoutput=1

\documentclass[11pt]{article}

\usepackage[]{ACL2023}

\usepackage{times}
\usepackage{latexsym}
\usepackage{graphicx}
\usepackage{tabularx}
\usepackage{makecell}
\usepackage{multirow}
\usepackage{hhline}
\usepackage{amsfonts}

\usepackage{booktabs}
\usepackage{enumitem}
\usepackage[T1]{fontenc}

\usepackage[utf8]{inputenc}

\usepackage{microtype}

\usepackage{inconsolata}

\title{A Simple and Effective Framework for Strict Zero-Shot Hierarchical Classification}
\author{Rohan Bhambhoria\thanks{\hspace{1mm}
This research was performed when the first author was a research intern at Rakuten Institute of Technology (RIT), Boston.}
{\hspace{0.01mm}} \textsuperscript{\rm 1}, Lei Chen\textsuperscript{\rm 2}, Xiaodan Zhu\textsuperscript{\rm 1} \\
        \textsuperscript{\rm 1}Department of Electrical and Computer Engineering \& Ingenuity Labs Research Institute \\ Queen's University, Canada \\  \textsuperscript{\rm 2}Rakuten Institute of Technology (RIT) \\ Boston, MA\\
        \texttt{\{r.bhambhoria,xiaodan.zhu\}@queensu.ca} \\  
 \texttt{lei.a.chen@rakuten.com}}

\begin{document}
\maketitle

\begin{abstract}

In recent years, large language models (LLMs) have achieved strong performance on benchmark tasks, especially in zero or few-shot settings. However, these benchmarks often do not adequately address the challenges posed in the real-world, such as that of hierarchical classification. In order to address this challenge, we propose refactoring conventional tasks on hierarchical datasets into a more indicative long-tail prediction task.
We observe LLMs are more prone to failure in these cases.
To address these limitations, we propose the use of entailment-contradiction prediction in conjunction with LLMs, which allows for strong performance in a strict zero-shot setting. Importantly, our method does not require any parameter updates, a resource-intensive process and achieves strong performance across multiple datasets.



\end{abstract}

\section{Introduction}
Large language models (LLMs) with parameters in the order of billions \cite{brown_language_2020} have gained significant attention in recent years due to their strong performance on a wide range of natural language processing tasks. These models have achieved impressive results on benchmarks \cite{https://doi.org/10.48550/arxiv.2204.02311}, particularly in zero or few-shot settings, where they are able to generalize to new tasks and languages with little to no training data. There is, however a difficulty in tuning parameters of these large-scale models due to resource limitations. Additionally, the focus on benchmarks has led to the neglect of real-world challenges, such as that of hierarchical classification. As a result, the long-tail problem \cite{gzs-longtail} has been overlooked. This occurs when a vast number of rare classes occur in the presence of frequent classes for many natural language problems.

\begin{figure}[!t]
    \centering \includegraphics[scale=0.32]{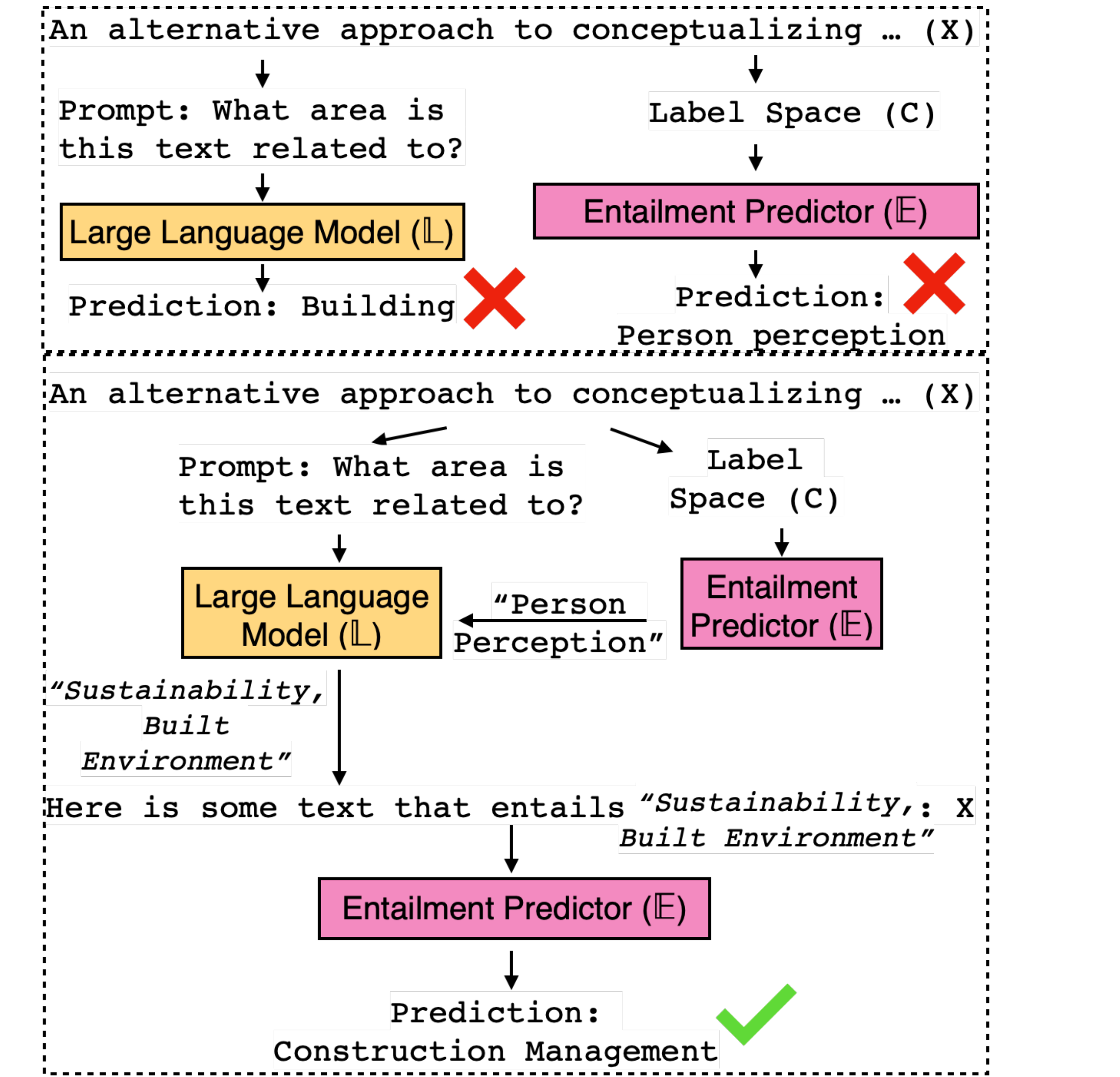}
    \caption{
    LLMs ($\mathbb{L}$) without any constraints and Entailment Predictors ($\mathbb{E}$) without guided knowledge (\textbf{Top}) show poor results independently. Our method (\textbf{Bottom}), combines advantages of these two systems to improve performance on strict zero-shot classification.}
    \label{fig:flow_diagram}
\end{figure}

In many industrial real-world applications, a strong performing method for hierarchical classification can be of direct utility. New product categories are emerging in e-commerce platforms. Existing categories, on the other hand, may not be very intuitive for customers. For example, upon browsing categories such as \textit{night creams}, we may be unable to find a product in a sibling-node category of \textit{creams}. This is further highlighted by platforms in which a systematic structure is not created for users; parent nodes may be in place of child nodes, and vice versa \cite{DBLP:journals/corr/Asghar16}. Manually categorizing product categories can be a costly redesigning endeavour. To tackle this problem, we suggest refactoring traditional hierarchical flat-labeled prediction tasks \cite{liu_improving_2021} to a more indicative long-tail prediction task. This involves structuring the classification task to closely reflect the real-world long-tail distribution of classes. In doing so, we are enabled to leverage LLMs for long-tail prediction tasks in a strict zero-shot classification setting. Through a series of experiments, results in this work show that our proposed method is able to significantly improve the performance over the baseline in several datasets, and holds promise for addressing the long-tail problem in real-world applications. The contributions of this work can be summarized as follows:
\begin{itemize}[leftmargin=*]
    \item We refactor real-world hierarchical taxonomy datasets into long-tailed problems. In doing so, we create a strong testbed to evaluate ``strict zero-shot classification" with LLMs.
    \item We explore utilizing LLMs to enhance the capabilities of entailment-contradiction predictors for long-tail classification. This results in strong capabilities of performing model inference without resource-intensive parameter updates.
    \item We show through quantitative empirical evidence, that our proposed method is able to overcome limitations of stand-alone large language models. Our method obtains strong performance on long-tail classification tasks.
\end{itemize}

\section{Background and Related Work}
\noindent\textbf{Strict Zero-Shot Classification} 

\noindent Previous works \cite{liu_improving_2021,yin_benchmarking_2019} have explored zero-shot classification extensively under two settings\textemdash(i) zero-shot, where testing labels are unseen, i.e. no overlap with the training labels, and (ii) generalized zero-shot, testing labels are partially unseen. In both cases, the model is trained on data from the same distribution as the test data. In our proposed \emph{strict} zero-shot setting, the model is only trained to learn the entailment relationship from natural language inference (NLI) corpora \cite{williams2018broad}. The training data for this model has no overlap with the distribution or semantics of the inference set. Additionally, previous works utilizing NLI have either not examined the utility of LLMs \cite{ye2020zero,gera_zero-shot_2022}, or transfer the capabilities of LLMs to smaller models but have failed to use them in a strict zero-shot setting for long-tail problems, only demonstrating their utility for benchmark tasks \cite{tam_improving_2021,schick_its_2021}. Works exploring LLMs have also limited their study to only using them independently without exploring entailment-contradiction prediction \cite{cot-prompting,brown_language_2020}.\newline


\begin{figure}
    \centering
    \includegraphics[scale=0.29]{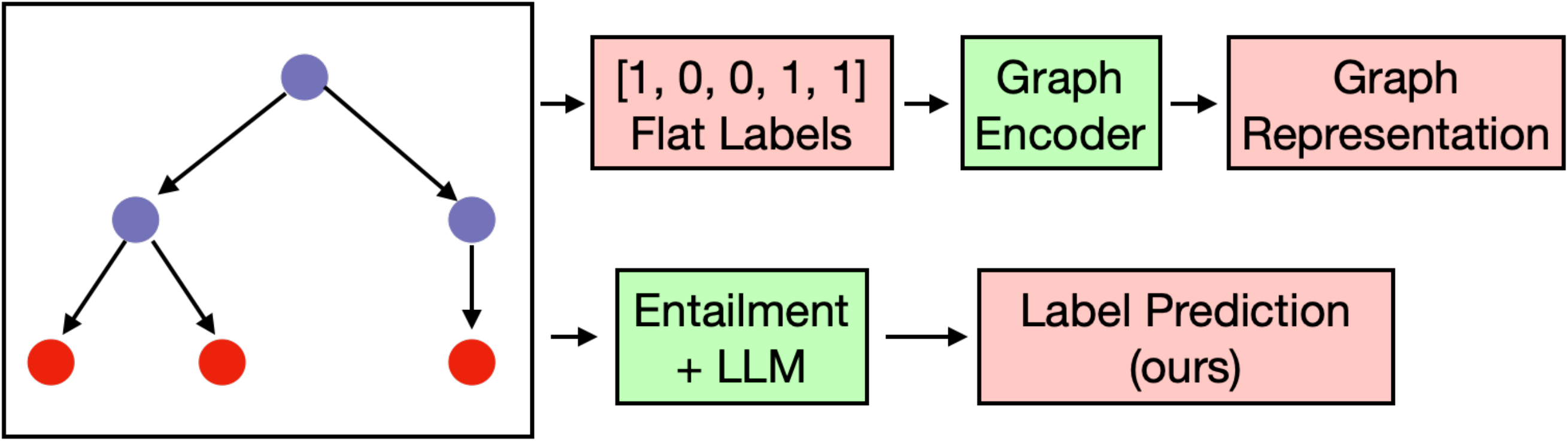}
    \caption{Previous flat label prediction and graph representation tasks (\textbf{Top}) do not make use of natural entailment relations. Our method for Label Prediction of red leaf nodes (\textbf{Bottom}) enables inference in a \textit{strict} zero-shot setting. 
    }
    \label{fig:label_prediction}
\end{figure}

\begin{figure*}[!t]
  \centering
  \begin{tabular}{ccc}
    \includegraphics[width=0.29\textwidth]{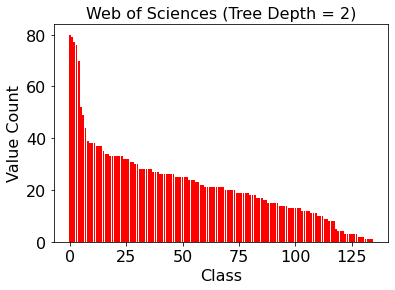} &
    \includegraphics[width=0.31\textwidth]{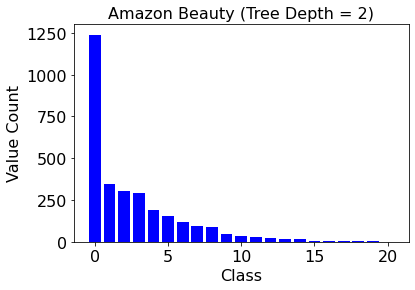} &
    \includegraphics[width=0.31\textwidth]{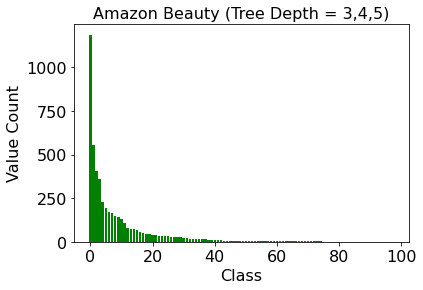} \\
  \end{tabular}
  \caption{Web of Science (WOS) and Amazon Beauty datasets, refactored to a long-tail distribution. Maximum tree depth is shown for Amazon Beauty, which varies from 3-5. Leaf nodes are used in our method regardless of depth.}
  \label{fig:longtail-distribution}
\end{figure*}

\noindent\textbf{Long Tail Problem}

\noindent
\citet{gzs-longtail,zhang_making_2022} highlight the significance of addressing the long-tail task. Existing literature in natural language processing has focused on scenarios involving limited data availability, such as few-shot or low-resource settings. It has failed to adequately address the unique challenges presented by long-tail problems. These problems arise when a small number of classes possess a large number of samples, while a large number of classes contain very few samples. Previous works have not delved into the specific use of LLMs or entailment predictors. \newline

\noindent\textbf{Hierarchical Classification}

\noindent Many real-world problems contain taxonomy data structured in a hierarchical setting. Shown in Figure \ref{fig:label_prediction}, most previous works make use of this data as a flat-label task \cite{kowsari2017HDLTex,zhou_hierarchy-aware_2020}. It is however, non-trivial to create clean training data for taxonomies, which these methods rely on.
This setting also combines parent and child nodes into a multi-label task, thereby increasing the complexity of the problem as siblings amongst leaf nodes are more diverse than parent nodes. Additionally, previous works do not make use of the natural entailment relations in hierarchies. Other works extenuate this problem by opting to utilize flat labels to produce graph representations \cite{wang_incorporating_2022,https://doi.org/10.48550/arxiv.2204.13413,jiang_exploiting_2022,chen_hierarchy-aware_2021}. For this reason, the graph representations may have limited value independently, although representations may be used to assist text classification by providing an organized label space. These works only introduce hierarchies to bring order to the label space, but overlook the original task of hierarchical taxonomy classification \cite{kowsari2017HDLTex}. For all previous works, difficult to obtain fine-tuning data is required to produce strong signals.
In our work, we refactor this data into a leaf-node label prediction task with the help of entailment-contradiction relations and LLMs. In doing so, we enable hierarchical taxonomy prediction independent of utilizing training data for the downstream task.


\section{Methodology}
\label{sec:method}
In this paper, we investigate the limitations of LLMs in three overlooked settings, when\textemdash (i) the model is not provided with sufficient examples due to the input text length, (ii) the label space includes tokens largely unobserved in the model's pretrained vocabulary, and (iii) there are a large number of distractors in the label space \cite{kojima_large_2022,min2022rethinking,razeghi_impact_2022}. These scenarios are common in real-world tasks, but are often overlooked in the development and evaluation of LLMs. To address these challenges, we propose the use of entailment-contradiction prediction \cite{yin_benchmarking_2019}, 
the task of determining whether a premise logically entails or contradicts a hypothesis. Through our method, we are able to leverage strong reasoning from \citet{yin_benchmarking_2019} with the retrieval abilities of LLMs \cite{DBLP:journals/corr/abs-2010-11967} to improve overall performance in a strict zero-shot setting, where the model must classify samples from a new task without any fine-tuning or additional examples used for training from the same distribution as the inference dataset. Importantly, our proposed combined method does not require parameter updates to the LLM, a resource-intensive process that is not always feasible with increasingly large model size \cite{https://doi.org/10.48550/arxiv.2204.02311}.

Our simple framework is shown in Figure \ref{fig:flow_diagram}. Considering the label space, $\mathcal{C} = \{C_{1}, C_{2},...C_{n}\}$ as the set of classes for any given dataset, and a text input, $X$, we can utilize the entailment predictor, $\mathbb{E}$ to make a \emph{contradiction}, or \emph{entailment} prediction on each label. This is done by using $X$ as the premise, and "This text is related to $C_{i}$." $\forall C_{i} \in \mathcal{C}$ as the hypothesis \cite{yin_benchmarking_2019}. In our work, the premise may be modified to include the prompt template. The prediction, $\mathbb{E}(X)$ lacks any external knowledge and is restricted to the label space, which may result in poor performance. $\mathbb{E}(X)$ can however, provide us with an implicit classification of the contradiction relation for sibling nodes. In our work, we use $\mathbb{E}(X)$ as an initializer for LLMs. We also regard it as a baseline as it shows strong performance independently. A LLM, $\mathbb{L}$ on the other hand, operates in an open space, with aforementioned shortcomings for classification tasks. For our purposes, we can regard this as a noisy knowledge graph \cite{DBLP:journals/corr/abs-2010-11967}, which may be utilized to predict ancestors or descendants of the target class. We consider the prediction made by the LLM as $\mathbb{L}(X)$. It is important to note that $\mathbb{L}(X)$ may or may not belong to $\mathcal{C}$.
We try several prompts for this purpose, shown in Appendix \ref{sec:prompt-templates}.

\begin{table*}
\centering
\renewcommand\arraystretch{0.91}
\resizebox{0.9\linewidth}{!}{
\begin{tabularx}{\textwidth}{c|XX|XX|XX}
    \toprule
  \makecell{ \textbf{Model}\\}& \multicolumn{2}{c|}{\makecell{\textbf{WOS} \\ \textbf{(Tree Depth = 2)}}} & \multicolumn{2}{c|}{\makecell{\textbf{Amzn Beauty} \\ \textbf{(Tree Depth = 2)}}} & \multicolumn{2}{c}{\makecell{\textbf{Amzn Beauty} \\ \textbf{(Tree Depth = 3, 4, 5)}}} \\ 
   & Acc. & Mac.F1 & Acc. & Mac.F1  & Acc. & Mac.F1  \\ \hline
     T0pp & 10.47  & 11.04  & 7.35  & 6.01  & 12.04  & 4.87   \\ 
     \makecell{BART-MNLI (Baseline)} & 61.09  & 68.93  & 60.80  & 51.15  & 41.68  &  49.35  \\ 
    \midrule
   \makecell{T0pp  + BART-MNLI} &  20.64 & 24.01  & 37.24  & 24.38  &  23.47 & 18.01   \\ 
  \makecell{BART-MNLI  + T0pp} & 60.40  & 68.92  &  58.79 & 51.94  &  \textbf{43.98} & 46.06  \\ 
  \makecell{BART-MNLI  + T0pp (Primed)} &  60.16 & 68.81  & 59.79  & \textbf{54.04 } & 39.10  & 46.50  \\ 
  \makecell{BART-MNLI  + T0pp (Primed+)} & \textbf{61.78}  & \textbf{69.48}  &  \textbf{64.25} & 52.84  & 40.79  & \textbf{49.96}  \\ 
  \bottomrule

\end{tabularx}}
\caption {Baseline models (\textbf{Top}) underperform our method (\textbf{Bottom}) across all datasets for average scores of Top-1, Top-3, and Top-5 accuracy and Macro F1. Our primed and primed+ models explicitly utilize the entailment relation, with one and five predictions of $\mathbb{L}(\mathbb{E}(X))$ respectively. All models used are available on Huggingface.}.
\label{avg-scores}
\end{table*}

By combining these two components, we can create a template which utilizes the \emph{entailment} relation explicitly, and the \emph{contradiction} relation implicitly
by constructing $\mathbb{L}(\mathbb{E}(X))$ to deseminate combined information into an entailment predictor for classification. The template we use is task-dependent, and is generally robust given an understanding of the domain. On Web of Sciences we use: "Here is some text that entails $\mathbb{E}(X)$: $X$. What area is this text related to?". For Amazon Beauty, we use "Here is a review that entails $\mathbb{E}(X)$: $X$. What product category is this review related to?". In this setting, our method still meets a barrier due to limitations of LLMs. By constructing a composite function, $\mathbb{E}(\mathbb{L}(\mathbb{E}(X))$, we are able to leverage our LLM in producing tokens which may steer the entailment predictor to correct its prediction. The template used for this composite function is "Here is some text that entails $\mathbb{L}(\mathbb{E}(X))$: $X$." across all datasets. 

\noindent\textbf{General Form:} Although our results are reported combining the advantages of $\mathbb{L}$, and $\mathbb{E}$ to produce upto the composite function $\mathbb{E}(\mathbb{L}(\mathbb{E}(X))$, this can be extended as it holds the property of being an iterative composition function to $\mathbb{E}(\mathbb{L}(\mathbb{E}(\mathbb{L}...\mathbb{E}(X))))$. Our observations show this setting having comparable, or marginal improvements with our dataset. However, this may prove to be beneficial in other tasks. We will investigate, and urge other researchers to explore this direction in future work.


\begin{figure}[t]
    \centering
    \includegraphics[scale=0.5]{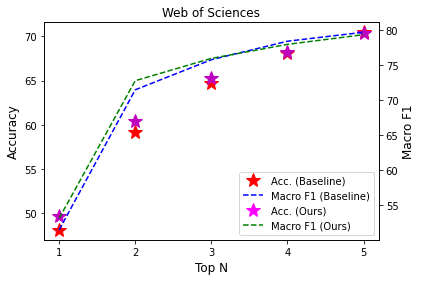}
    \caption{Results for Top-5 predictions on WOS dataset. BART-MNLI + T0pp (Primed+) (\textbf{ours}) converges with performance of the BART-MNLI (\textbf{baseline}) at Top-4.}
    \label{fig:WOS_results}
\end{figure}

\section{Experiments and Results}
\subsection{Dataset and Experimental Settings}
We refactor the widely used Web of Sciences (WOS) with \citet{kowsari2017HDLTex}, and Amazon Beauty \cite{10.1145/2766462.2767755} datasets to follow a class-wise long-tail distribution as shown in Figure \ref{fig:longtail-distribution}. Additionally, we create two variations of the Amazon Beauty dataset, first in which it contains the same tree depth as WOS, both containing 3000 samples, and second in which all classes are included for their maximum tree depth, containing 5000 samples. 
We select these datasets as they challenge the shortcomings of LLMs. The input text of providing multiple abstracts in the WOS dataset surpasses the maximum input token length of most transformer-based models. This makes it difficult for models to learn the input distribution, a requirement for showing strong in-context performance \cite{min2022rethinking}. Next, many tokens in the label space of both the WOS and Amazon Beauty datasets rarely occur in pretraining corpora, details of which are provided in the Appendix \ref{sec:stats}. Additionally, both datasets contain a large number of distractors, or incorrect classes in the label space. Further details are provided in Appendix \ref{sec:det-results}.

All experiments are performed on a single NIVIDA Titan RTX GPU. We use BART-Large-MNLI with 407M parameters as our baseline. We use this model as it outperforms other architectures trained on MNLI for zero-shot classification. For our LLM, we opt to use T0pp \cite{sanh_multitask_2022} with 11B parameters\footnote{We observe a significant drop in performance when we utilize the 3B parameter variant of this model as $\mathbb{L}$.}, as previous works show that it matches or exceeds performance of other LLMs such as GPT-3 \cite{brown_language_2020} on benchmarks. 

\subsection{Results and Discussion}
Results of our method are shown in Table \ref{avg-scores}. LLMs, due to their limitations, perform poorly as a standalone model for long-tail classification. These results can be improved by priming the model with an entailment predictor through the usage of a prompt. The baseline shows strong performance independent of the LLM, as it operates on a closed label space. The capabilities of the baseline can be enhanced by further explicitly priming it with a entailment relation through a LLM. Rows in which T0pp is initialized, or primed with $\mathbb{E}$ are indicated with \emph{Primed}. Priming the model showcases improvements across all datasets for macro F1. For accuracy, priming the model shows benefit in two out of three datasets.
In Figure \ref{fig:WOS_results}, we show the results of Top-5 predictions for the WOS dataset. All results are aggregated in Table \ref{avg-scores}. It is important to highlight that prompt variation led to stable results for our LLM. The variance upon utilizing BART-MNLI is negligible across prompts. The best results are observed upto Top-4 predictions on both accuracy and macro F1 for our method, when the entailment prompt is enhanced with a greater number of tokens corresponding to the output of $\mathbb{L}(\mathbb{E}(X))$. The variation between our method and the baseline is much greater for Top-1 predictions, but Top-5 prediction variance is negligible. Detailed results for both depth settings of Amazon Beauty are shown in Appendix \ref{sec:det-results}.
\section{Conclusion}
In this work, we utilize an LLM in the form of a noisy knowledge graph to enhance the capabilties of an entailment predictor. In doing so, we achieve strong performance in a strict zero-shot setting on several hierarchical prediction tasks. We also show the necessity of refactoring existing hierarchical tasks into long-tail problems that may be more representative of the underlying task itself. The utility in practical industry settings is also highlighted through this setting.

\section*{Limitations}
In this work, we implicitly utilize the \textit{contradiction} relation. The authors recognize explicitly including it in a prompt template leads to worse performance due to the injection of noise. Controlled template generation based on a model confidence is unexplored in this work and appears to be a promising direction. Additionally, we recognize the emergence of parameter-efficient methods for training models which are unexplored in this work, which may have utility. These methods are complimentary and may benefit the performance of models as they can be used in conjunction with training paradigms such as contrastive learning to support better representations through explicit utilization of the \emph{contradiction} relation. In this work, we limit our study to draw attention to the importance of strict zero-shot  classification settings with the emergence of LLMs. 

Our study can be easily extended to recursively operate on large language models, and entailment predictors. As we observe limited performance benefits in doing so, we conduct our study to show improvements after one complete cycle, given by $\mathbb{E}(\mathbb{L}(\mathbb{E}(X))$ in Section \ref{sec:method}.

\section*{Ethics Statement}

In this work, we propose a framework which allows for the usage of entailment-contradiction predictors in conjunction with large language models. In doing so, this framework operates in a stict zero-shot setting. While it is possible to tune prompts to select optimal variants through hard/soft prompt tuning strategies, this would require additional computational resources for LLMs with billions of parameters. Our study investigates the usage of LLMs given an understanding of the domain they tend to be used for (e.g., given an understanding of Amazon Beauty containing reviews, a prompt is constructed). Further explanation of prompt templates is contained in Appendix \ref{sec:prompt-templates}. Due to the lack of tuning parameters in this work, large language models are largely dependent on pre-training data. Although this can be controlled to some degree by introducing an entailment predictor with a fixed label space, the underlying model does not explicitly contain supervision signals without further training. The framework proposed for inference in this work must hence be used cautiously for sensitive areas and topics.


\bibliography{anthology,custom}
\bibliographystyle{acl_natbib}

\appendix

\begin{table*}
\centering
\renewcommand\arraystretch{0.91}
\resizebox{0.8\linewidth}{!}{
\begin{tabular}{c|c}
\toprule
\textbf{Dataset}                        & \textbf{Prompt} \\ \hline
\multirow{2}{*}{WOS}           &   What field is this passage related to? + X     \\
                               &     What area is this text related to? + X   \\ 
                               &     X + What area is this text related to?  \\ 
                               &     What area is this text related to? + X   \\ 
                               \midrule
\multirow{2}{*}{Amazon Beauty} &    Here is a review: + X + What product category is this review related to?    \\
                               & X + What product category is this text related to?  \\ \bottomrule

\end{tabular}}
\caption{Example prompts used to initialize the LLM, $L$. }
\label{tab:prompt-guide}
\end{table*}

\section{Prompt Templates}
\label{sec:prompt-templates}

In our work, we try various prompts for WOS and Amazon Beauty to initialize the LLM, and for the entailment predictor. These prompts are shown in Table \ref{tab:prompt-guide}. Initializing prompts for $\mathbb{L}$ may show some variance in performance when utilized independently. The prompts used for obtaining $\mathbb{L}(\mathbb{E}(X))$ are generally robust with an understanding of the domain, and show a marginal impact on outcome, upon variation. Prompts used for $\mathbb{E}(\mathbb{L}(\mathbb{E}(X)))$ have an insignificant impact on the outcome.

\section{Statistics}
\label{sec:stats}

We provide some details of the distribution for Web of Science dataset are provided with the head, and tail of the distribution class names with their respective value count in Table \ref{tab:wos-dist}. We also provide the details of class-wise distribution for Amazon Beauty (Depth=2), and Amazon Beauty (Depth=3,4,5) datasets in Table \ref{tab:amzn2-dist}, and Table \ref{tab:amzn3-dist} respectively. Towards the tail-end of the distribution, we observe several tokens which may infrequently appear in most pretraining corpora, such as "Polycythemia Vera" for the WOS dataset. Updating parameters of a model on data which is heavily skewed towards the tail distribution in the presence of frequently occuring labels can be problematic for language models. Our proposed method in this work is one solution towards this challenging task.

\section{Detailed Results}
\label{sec:det-results}

We provide some details of results for Top-1, Top-3, and Top-5 accuracies and macro F1 scores in this section. The Web of Sciences dataset results are shown in Table \ref{tab:wos-details}. We observe that the accuracy is significantly higher by all of our methods over the baseline, BART-MNLI. The same trends are seen for Macro F1 scores. In predicting Top-3 labels, only our method of Primed+ shows improvement over the baseline. For macro F1, our method in the Top-3 category shows slight improvement over the baseline. For Top-5 predictions on the WOS dataset, our method shows performance marginally below the baseline. Results for Amazon Beauty (Depth=2) are shown in Table \ref{tab:amzn2-details}. There is a large improvement in accuracy using our method on this dataset for Top-1, 3, and 5. For Macro F1, there our method performs marginally worse than the baseline for Top-1 predictions. Our method strongly outperforms the baseline by a large margin for Top-3 and Top-3 prediction on Macro F1. The results for Amazon Beauty (Depth=3,4,5) are shown in Table \ref{tab:amzn3-details}. Our method improves upon the baseline for both, accuracy and macro F1 for Top-1 predictions. For Top-3, our method has a significant improvement over accuracy, with comparable performance on Macro F1. Our method has a large improvement on Top-5 scores for accuracy, and improves upon the Macro F1 score for Macro F1.

With our dataset settings, we observe the performance of using int-8 quantization is robust and matches that of bf-16/fp-32 for inference. These settings also provide us with stable results across prompts. 

Previous works have performed parameter-updates \cite{gera_zero-shot_2022,holtzman_surface_2021} to models to tackle the challenge of many distractors in the label space. This may be practically infeasible due to the requirements of compute in the case of LLMs. 

Diversity between category labels is an important factor we observe which attributes to the improvement in performance. Tables \ref{tab:wos-dist}, \ref{tab:amzn2-dist}, \ref{tab:amzn3-dist} contain statistics for labels used. We observed a significant drop in Macro F1 shown in Table \ref{avg-scores} for the Amazon Beauty Dataset (Tree Depth=2) in contrast to WOS for the same models due to the lack of diversity in several class names (e.g. “Bath” and “Bathing Accessories”). Similar trends were observed in Amazon Beauty (Tree Depth=3,4,5) for “Eau de Toilette” and “Eau de Parfum”, both of which are perfumes.

\begin{table}[!h]
    \centering
    \begin{tabular}{c|c}
    \toprule
         \textbf{Class Name} & \textbf{Value Count} \\
         \hline
        Polymerase chain reaction & 95 \\ 
        Northern blotting & 88 \\
        Molecular biology & 66 \\
        Human Metabolism  & 65 \\
        Genetics & 62 \\
        Stealth Technology  & 2\\
        Voltage law   & 1 \\
        Healthy Sleep & 1 \\
        Kidney Health & 1 \\
        Polycythemia Vera  & 1 \\
    \bottomrule
    \end{tabular}
    \caption{Class names and corresponding value counts for head and tail elements from for WOS dataset.}
    \label{tab:wos-dist}
\end{table}

\begin{table}[!t]
    \centering
    \begin{tabular}{c|c}
    \toprule
         \textbf{Class Name} & \textbf{Value Count} \\
         \hline
        Face & 1230 \\ 
        Body & 344 \\
        Styling Products & 298 \\
        Women's & 289 \\
        Styling Tools & 187 \\
        Bags \& Cases & 5\\
        Hair Loss Products  & 5 \\
        Bath & 3 \\
        Bathing Accessories & 2 \\
        Makeup Remover  & 1 \\
    \bottomrule
    \end{tabular}
    \caption{Class names and corresponding value counts for head and tail elements from Amazon Beauty (Depth=2) dataset.}
    \label{tab:amzn2-dist}
\end{table}

\begin{table}[!t]
    \centering
    \begin{tabular}{c|c}
    \toprule
         \textbf{Class Name} & \textbf{Value Count} \\
         \hline
        Lotions & 1188 \\ 
        Eau de Toilette & 553 \\
        Nail Polish & 405 \\
        Eau de Parfum & 363 \\
        Soaps & 231 \\
        Shower Caps & 1\\
        Paraffin Baths  & 1 \\
        Hairpieces & 1 \\
        Tote Bags & 1 \\
        Curlers  & 1 \\
    \bottomrule
    \end{tabular}
    \caption{Class names and corresponding value counts for head and tail elements from Amazon Beauty (Depth=3,4,5) dataset.}
    \label{tab:amzn3-dist}
\end{table}

\begin{table*}[t!]
\centering
\renewcommand\arraystretch{0.91}
\resizebox{0.9\linewidth}{!}{
\begin{tabularx}{\textwidth}{c|XX|XX|XX}
\toprule
  \makecell{ \textbf{Model}\\}& \multicolumn{2}{c|}{\textbf{Top-1}} & \multicolumn{2}{c|}{\makecell{\textbf{Top-3} \\ }} & \multicolumn{2}{c}{\makecell{\textbf{Top-5} \\ }} \\ 
  &
  Acc. &
  Mac. F1 &
  Acc. &
  Mac. F1 &
  Acc. &
  Mac. F1 \\ \hline
T0pp & 5.46  & 5.66  & 11.26 & 12.25 & 14.70  & 15.23 \\ 
BART-MNLI & 48.10  & 51.49 & \textbf{64.73} & \textbf{75.77} & \underline{\textbf{70.46}} & \underline{\textbf{79.69}}  \\
 \midrule
\makecell{T0pp  + BART-MNLI} & 12.10  & 13.75 & 22.3  & 26.44 & 27.53 & 31.84  \\

 \makecell{BART-MNLI  + T0pp} & 48.16 & 52.16 & 63.80 & 75.40 & 69.26 & 79.20\\
\makecell{BART-MNLI  + T0pp (Primed)} & \textbf{48.69} & \textbf{52.78} & 63.60  & 75.29 & 68.20  & 78.37  \\ 

 \makecell{BART-MNLI  + T0pp (Primed+)} & \underline{\textbf{49.73}} & \underline{\textbf{53.15}} & \underline{\textbf{65.23}} & \underline{\textbf{75.96}}& \textbf{70.39} & \textbf{79.34} \\ \bottomrule
\end{tabularx}}
\caption{Accuracy and Macro F1 results for Top-1, Top-3, and Top-5 predictions for the Web of Sciences dataset.}
\label{tab:wos-details}
\end{table*}

\begin{table*}
\centering
\renewcommand\arraystretch{0.91}
\resizebox{0.9\linewidth}{!}{
\begin{tabularx}{\textwidth}{c|XX|XX|XX}
\toprule
 \makecell{ \textbf{Model}\\}& \multicolumn{2}{c|}{\textbf{Top-1}} & \multicolumn{2}{c|}{\makecell{\textbf{Top-3} \\ }} & \multicolumn{2}{c}{\makecell{\textbf{Top-5} \\ }} \\ 
  &
  Acc. &
  Mac. F1 &
  Acc. &
  Mac. F1 &
  Acc. &
  Mac. F1 \\ \hline
T0pp & 3.99  & 2.58  & 7.48 & 7.08 & 10.57  & 8.37 \\ 
BART-MNLI & \textbf{34.40}  & \underline{\textbf{25.10}} & \textbf{68.54} & \textbf{60.15} & 79.45 & 68.21  \\ \midrule
\makecell{T0pp + BART-MNLI} & 19.87  & 8.95 & 39.94  & 26.30 & 51.93 & 37.89  \\ 

 \makecell{BART-MNLI + T0pp} & 33.36 & \textbf{24.84} & 61.12 & 58.63 & \textbf{81.90} & 72.34\\ 
\makecell{BART-MNLI + T0pp (Primed)} & \underline{\textbf{41.22}} & 24.30 & 61.46  & 60.22 & 76.70  & \underline{\textbf{77.59}}  \\ 
 \makecell{BART-MNLI + T0pp (Primed+)} & 32.32 & 19.91 & \underline{\textbf{75.19}} & \underline{\textbf{63.74}}& \underline{\textbf{85.26}} & \textbf{74.87} \\ \bottomrule
\end{tabularx}}
\caption{Accuracy and Macro F1 results for Top-1, Top-3, and Top-5 predictions for the Amazon Beauty dataset (depth = 2).}
\label{tab:amzn2-details}
\end{table*}

\begin{table*}
\centering
\renewcommand\arraystretch{0.91}
\resizebox{0.9\linewidth}{!}{
\begin{tabularx}{\textwidth}{c|XX|XX|XX}
\toprule
 \makecell{ \textbf{Model}\\}& \multicolumn{2}{c|}{\textbf{Top-1}} & \multicolumn{2}{c|}{\makecell{\textbf{Top-3} \\ }} & \multicolumn{2}{c}{\makecell{\textbf{Top-5} \\ }} \\ 
  &
  Acc. &
  Mac. F1 &
  Acc. &
  Mac. F1 &
  Acc. &
  Mac. F1 \\ \hline
T0pp & 5.22  & 2.32  & 13.80 &  5.54  & 17.12 & 6.76\\ 
BART-MNLI & \textbf{32.58}  & \textbf{28.05} & 43.73 & \underline{\textbf{56.18}} & 48.75 & 63.83  \\ \midrule
\makecell{T0pp + BART-MNLI} & 12.49  & 6.99 & 26.26  & 20.64 & 31.67 & 26.41  \\ 

 \makecell{BART-MNLI + T0pp} & \underline{\textbf{33.89}} & 23.15 & \underline{\textbf{47.06}} & 53.02 & \textbf{51.01} & 62.02\\ 
\makecell{BART-MNLI + T0pp (Primed)} & 28.18 & 20.22 & 41.89  & 55.15 & 47.24  & \underline{\textbf{64.14}}  \\

 \makecell{BART-MNLI + T0pp (Primed+)} & 23.92 & \underline{\textbf{29.70}} & \textbf{46.43} & \textbf{56.07}& \underline{\textbf{52.02}} & \textbf{64.11} \\ \bottomrule
\end{tabularx}}
\caption{Accuracy and Macro F1 results for Top-1, Top-3, and Top-5 predictions for the Amazon Beauty dataset (depth = 3,4,5).}
\label{tab:amzn3-details}
\end{table*}

\end{document}